\documentclass[preprint,12pt]{article}

\usepackage[utf8]{inputenc}
\usepackage[T1]{fontenc}
\usepackage{amsmath,amssymb}
\usepackage{graphicx}
\usepackage{booktabs}
\usepackage{hyperref}
\usepackage{url}
\usepackage{textcomp}
\usepackage{multirow}
\usepackage{array}
\usepackage{authblk}
\usepackage[margin=1in]{geometry}
\usepackage{lineno}
\usepackage{caption}

\DeclareCaptionLabelFormat{boldlabel}{\textbf{#1 #2.}}
\begin{document}

\title{\bfseries Adversarial Robustness of Deep Learning-Based Thyroid Nodule Segmentation in Ultrasound}

\author[1]{Nicholas Dietrich}
\author[2]{David McShannon}

{\tiny
\affil[1]{Temerty Faculty of Medicine, University of Toronto, Toronto, Ontario, Canada}
\affil[2]{Faculty of Engineering, McMaster University, Hamilton, Ontario, Canada}
}

\date{}

\maketitle

\begin{abstract}
\noindent\textit{Introduction:} Deep learning-based segmentation models are increasingly integrated into clinical imaging workflows, yet their robustness to adversarial perturbations remains incompletely characterized, particularly for ultrasound images. We evaluated adversarial attacks and inference-time defenses for thyroid nodule segmentation in B-mode ultrasound. \textit{Methods:} Two black-box adversarial attacks were developed: (1) Structured Speckle Amplification Attack (SSAA), which injects boundary-targeted noise, and (2) Frequency-Domain Ultrasound Attack (FDUA), which applies bandpass-filtered phase perturbations in the Fourier domain. Three inference-time mitigations were evaluated on adversarial images: randomized preprocessing with test-time augmentation, deterministic input denoising, and stochastic ensemble inference with consistency-aware aggregation. Experiments were conducted on a U-Net segmentation model trained on cine-clips from a database of 192 thyroid nodules. \textit{Results:} The baseline model achieved a mean Dice similarity coefficient (DSC) of 0.76 (SD 0.20) on unperturbed images. SSAA reduced DSC by 0.29 (SD 0.20) while maintaining high visual similarity (SSIM = 0.94). FDUA resulted in a smaller DSC reduction of 0.11 (SD 0.09) with lower visual fidelity (SSIM = 0.82). Against SSAA, all three defenses significantly improved DSC after correction, with deterministic denoising showing the largest recovery (+0.10, p $<$ 0.001), followed by randomized preprocessing (+0.09, p $<$ 0.001), and stochastic ensemble inference (+0.08, p = 0.002). No defense achieved statistically significant improvement against FDUA. \textit{Conclusion:} Spatial-domain adversarial perturbations in ultrasound segmentation showed partial mitigation with input preprocessing, whereas frequency-domain perturbations were not mitigated by the defenses, highlighting modality-specific challenges in adversarial robustness evaluation.
\end{abstract}


\newpage
\section{Introduction}

Automated segmentation of thyroid nodules from B-mode ultrasound is a clinically important task that supports diagnosis, treatment planning, and longitudinal monitoring \cite{ref1,ref2}. In recent years, deep learning approaches have shown promising performance on such tasks, enabling accurate delineation of nodule boundaries across datasets \cite{ref2,ref3,ref4}. As these models move closer to clinical deployment, it becomes increasingly important to understand how and when they fail, particularly in settings that may compromise reliability or patient safety.

Adversarial attacks, which introduce small, often imperceptible perturbations to input images that can cause substantial failures in neural network predictions, have been widely studied in natural image classification \cite{ref5,ref6,ref7}. However, their implications for medical image segmentation remain comparatively underexplored \cite{ref5,ref6,ref8}. Ultrasound imaging presents several properties that are especially relevant to adversarial manipulation. Unlike natural photographs, ultrasound images are dominated by multiplicative speckle noise, exhibit characteristic frequency-domain structure arising from beam formation, and display high inter-frame variability that may obscure subtle perturbations \cite{ref9,ref10,ref11}. These characteristics suggest that ultrasound-specific attack strategies may differ meaningfully from generic perturbation methods and may be more difficult to detect visually.

Prior work on adversarial attacks in medical imaging has largely focused on classification tasks and has often assumed white-box access to model parameters and gradients \cite{ref5,ref6}. In contrast, black-box attacks, where the attacker has access only to model inputs and outputs, have received less attention, particularly for ultrasound segmentation \cite{ref6,ref12}. Existing studies involving ultrasound have primarily examined classification tasks or relied on generic pixel-level perturbations without explicitly leveraging ultrasound-specific image properties \cite{ref12,ref13}. Segmentation tasks introduce a broader attack surface than classification, as adversarial perturbations can degrade spatial accuracy, shift predicted boundaries, or suppress detected regions entirely \cite{ref6,ref14,ref15}.

Equally important to characterizing attacks is the evaluation of practical defenses. Adversarial training requires retraining models and access to representative attack distributions, while certified defenses based on randomized smoothing offer formal guarantees at the cost of reduced accuracy and increased computational burden \cite{ref5,ref16}. In contrast, inference-time defenses that preprocess inputs or aggregate stochastic predictions can be deployed without modifying the trained model and are compatible with existing architectures \cite{ref5,ref17}. To our knowledge, the effectiveness of such lightweight defenses against ultrasound-tailored adversarial attacks has not been evaluated.

In this study, we aimed to (1) train and evaluate a baseline deep learning model for thyroid nodule segmentation in ultrasound; (2) develop two adversarial attacks tailored to ultrasound imaging under realistic black-box constraints; and (3) evaluate three inference-time mitigation strategies against these attacks to assess their effectiveness.

\section{Materials and Methods}

\subsection*{\textit{Dataset}}

We used the publicly available Stanford AIMI Thyroid Ultrasound Cine-clip database, comprising 192 biopsy-confirmed thyroid nodules from 167 patients, with B-mode ultrasound cine-clips and radiologist-annotated nodule segmentation masks \cite{ref18}. Each cine-clip corresponds to distinct thyroid nodules and contains multiple sequential frames. All 17,412 frames were used without subsampling. Images were resized from 802 $\times$ 1054 pixels to 256 $\times$ 256 pixels and normalized to [0, 1]. The dataset was partitioned at the patient level (70/15/15 split) to prevent data leakage, yielding 140 nodules (12,889 frames) for training, 21 nodules (1,735 frames) for validation, and 31 nodules (2,788 frames) for testing.

\subsection*{\textit{Segmentation Model}}

The target segmentation model was a U-Net \cite{ref19} with an encoder of four stages (32, 64, 128, 256 features), a 512-feature bottleneck, and a symmetric decoder with skip connections. Batch normalization and ReLU activations were used throughout, with dropout (rate 0.1) during training. Training augmentations included random horizontal flips, vertical flips, and random 90\textdegree{} rotations. The model was trained with AdamW (lr = 1$\times$10$^{-4}$, weight decay = 1$\times$10$^{-5}$) using a combined binary cross-entropy and Dice loss, with a cosine annealing warm restarts scheduler (T$_0$ = 20, T\_mult = 2) and a batch size of 16. Training was performed on a single NVIDIA A100 80 GB GPU and completed in 26.4 minutes. Early stopping with patience of 15 epochs halted training at epoch 20, selecting the best model from epoch 5.

\subsection*{\textit{Threat Model}}

We adopted a black-box threat model in which the attacker can query the segmentation model with arbitrary input images and observe the predicted binary mask, but has no access to model weights, gradients, or architecture details. We intended for this set-up to reflect realistic deployment scenarios where models are accessed through inference APIs or embedded clinical devices. The attacker is permitted a fixed query budget of 500 per clip (50 iterations with a population size of 10). The attacker's objective was to minimize the Dice similarity coefficient (DSC) between the model's prediction on the perturbed image and the ground truth segmentation mask, while maintaining visual imperceptibility.

\subsection*{\textit{Attack Methods}}

\subsubsection*{Structured Speckle Amplification Attack}

SSAA aims to exploit the multiplicative speckle noise model inherent to ultrasound by injecting spatially structured noise concentrated near the predicted segmentation boundary (Figure 1). The attack first computes the Euclidean distance transform from the boundary of the model's current predicted mask. A Gaussian weighting function, with center offset randomly sampled from 0 to 15 pixels and spatial scale (sigma) sampled from 3 to 30 pixels, determines where perturbations are strongest. Rayleigh-distributed noise, which approximates the statistical distribution of ultrasound speckle, is generated, spatially smoothed with a Gaussian kernel (sigma = 1.5), and then zero-centered and normalized to unit variance. The perturbation is applied multiplicatively to the original image with an amplitude factor randomly sampled between 0.03 and 0.20, followed by clipping to the valid intensity range [0, 1]. All SSAA attacks operated under the query constraints defined in the threat model.

\begin{figure}[ht]
\centering
\includegraphics[width=\textwidth]{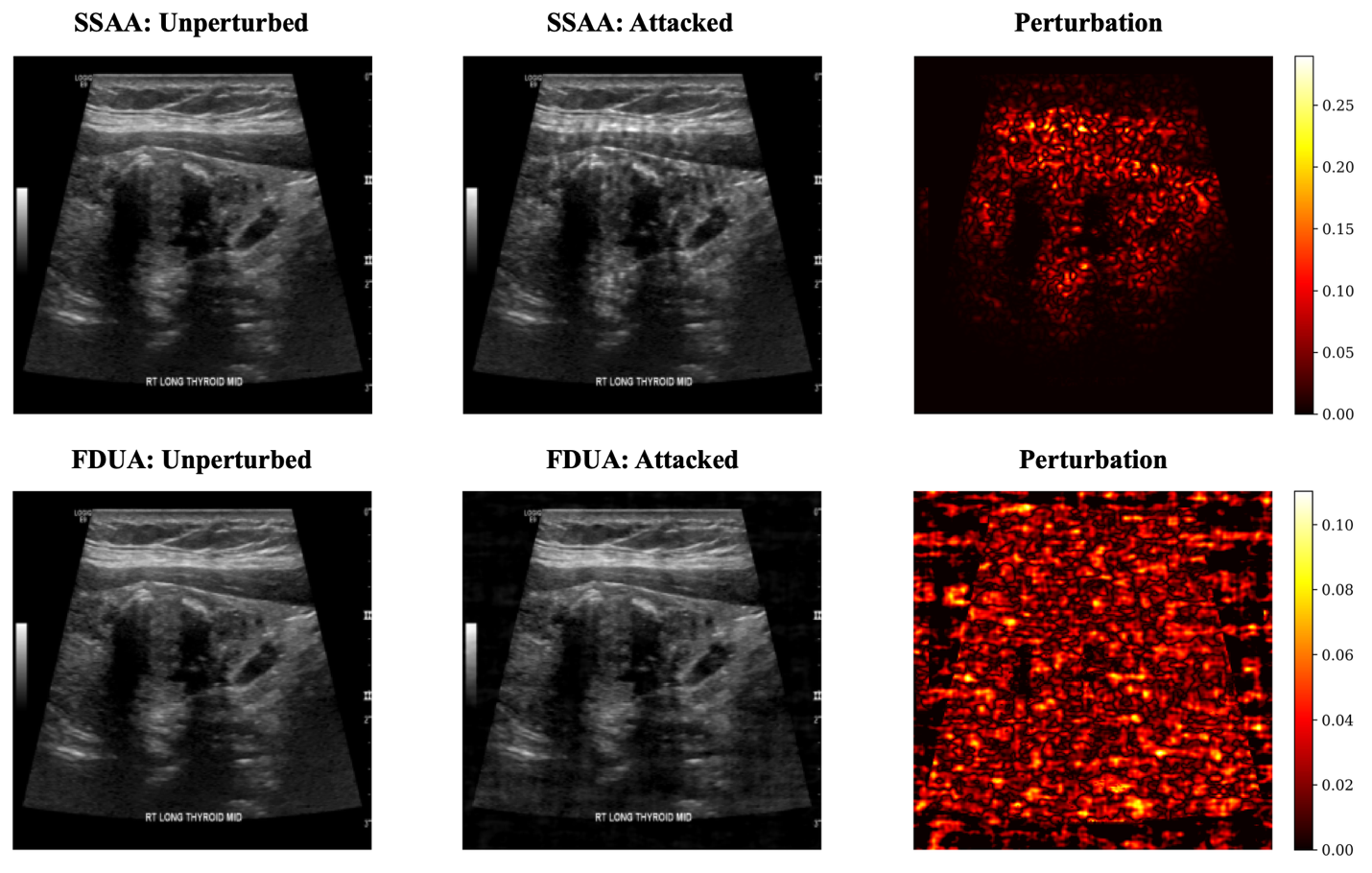}
\caption{Representative examples of adversarial perturbations applied to thyroid ultrasound images. Top row = Spatial Speckle-based Adversarial Attack (SSAA). Bottom row = Frequency-Domain Ultrasound Attack (FDUA). For each attack, the unperturbed input image, adversarially perturbed image, and corresponding perturbation magnitude map are shown.}
\label{fig:fig1}
\end{figure}

\subsubsection*{Frequency-Domain Ultrasound Attack}

FDUA aimed to introduce perturbations by modifying the 2D Fourier transform of the input image within a targeted frequency band (Figure 1). The image is transformed to the frequency domain and centered. A Butterworth bandpass filter isolated a target frequency range, with the low cutoff sampled uniformly from 5 to 38 cycles per image (approximately 4\% to 30\% of the Nyquist frequency), the high cutoff sampled from 38 to 102 cycles per image (approximately 30\% to 80\% of the Nyquist frequency), and filter order randomly chosen from 1 to 4. Random phase noise uniformly distributed over [$-\pi$, $\pi$] is generated, scaled by an amplitude parameter (epsilon sampled between 0.05 and 0.50), and applied as a complex multiplicative perturbation to the bandpass-filtered frequency components. The perturbed representation is inverse-transformed to the spatial domain and clipped to valid intensity values. This approach targeted mid-frequency components where tissue texture information is concentrated. All FDUA attacks also operated under the query constraints defined in the threat model.

\subsection*{\textit{Defense Strategies}}

Adversarial images were generated once against the undefended model and then evaluated through each defense independently. The attacker does not adapt to the presence of defenses, which reflects a realistic scenario where defenses are deployed after model training without the attacker's knowledge. This non-adaptive evaluation is a standard protocol in the black-box defense literature \cite{ref5,ref6}.

\subsubsection*{Randomized Preprocessing with Test-Time Augmentation}

This defense applies K = 5 random transformations to the input before inference and averages the resulting probability maps. Each transformation consists of a random spatial rescaling (factor sampled uniformly between 0.9 and 1.1, followed by center cropping or padding to restore the original size) and a random Gaussian blur (sigma sampled uniformly between 0.3 and 1.5). The averaged probability map is thresholded at 0.5 to produce the final binary prediction.

\subsubsection*{Deterministic Input Denoising}

This defense applies a fixed preprocessing pipeline consisting of a Gaussian blur with sigma = 1.0 followed by a 3 by 3 median filter before standard model inference. This removes high-frequency perturbation components while preserving coarser structural information needed for segmentation. As the only defense without stochastic components, it is deterministic and requires only a single forward pass.

\subsubsection*{Stochastic Ensemble with Consistency-Aware Aggregation}

This defense generates K = 5 augmented copies of the input, each subjected to random spatial shifts (uniformly sampled between $-$4 and +4 pixels with reflective boundary padding), random rescaling (factor between 0.93 and 1.07), random Gaussian blur (sigma between 0.2 and 1.2), random additive Gaussian noise (standard deviation between 0.005 and 0.02), and random brightness shift (between $-$0.03 and +0.03). Each copy is passed through the model to produce a probability map. A majority vote across the K predictions determines a preliminary consensus mask. Each prediction is then weighted by its pixel-wise agreement with the majority, such that pixels with unanimous agreement receive high weight and pixels with disagreement receive low weight. The final prediction is the weighted average of all probability maps, thresholded at 0.5.

\subsection*{\textit{Evaluation Metrics}}

Segmentation quality was assessed using the DSC, intersection-over-union (IoU), precision, recall, pixel accuracy, and the 95th percentile Hausdorff distance (HD95). Imperceptibility of adversarial perturbations was quantified using the structural similarity index (SSIM), L$_2$ norm, and L-infinity norm. For each mitigation, the recovery rate was computed as the fraction of the attack-induced Dice loss restored by the defense: (mitigated Dice minus undefended Dice) divided by (unperturbed Dice minus undefended Dice). The unperturbed performance cost of each defense was measured by evaluating it on unattacked images.

\subsection*{\textit{Statistical Analysis}}

All comparisons between defended and undefended conditions used a paired design, as the same 31 adversarial images (one per test nodule) were evaluated under each condition. The one-sided Wilcoxon signed-rank test was used to assess whether each defense significantly improved the adversarial DSC. The DSC was designated as the primary metric, with six comparisons (three defenses by two attacks) corrected using the Bonferroni method. IoU, precision, recall, and pixel accuracy were designated as secondary metrics. Effect sizes are reported as Cohen's d computed on paired differences, with interpretation thresholds of less than 0.2 (negligible), 0.2 to 0.5 (small), 0.5 to 0.8 (medium), and 0.8 or greater (large). Bootstrap 95\% confidence intervals (1,000 resamples) are reported for all DSC improvement estimates.

\section{Results}

\subsection*{\textit{Baseline Model Performance}}

On the 31 test cine clips, the U-Net model achieved a mean DSC of 0.76 (SD 0.20), IoU of 0.65 (SD 0.23), precision of 0.81 (SD 0.25), recall of 0.76 (SD 0.19), pixel accuracy of 0.95 (SD 0.04), and HD95 of 32.88 (SD 26.88) pixels. Table 1 provides the summary of segmentation metrics across all conditions.

\begin{table}[ht]
\tiny
\begin{tabular}{lcccccc}
\toprule
\textbf{Condition} & \textbf{DSC} & \textbf{IoU} & \textbf{Precision} & \textbf{Recall} & \textbf{Pixel Accuracy} & \textbf{HD95} \\
\midrule
Unperturbed (baseline) & 0.76 (0.20) & 0.65 (0.23) & 0.81 (0.25) & 0.76 (0.19) & 0.95 (0.04) & 32.88 (26.88) \\
Unperturbed + Randomized Preprocessing & 0.76 (0.20) & 0.65 (0.23) & 0.78 (0.25) & 0.79 (0.17) & 0.95 (0.04) & 37.05 (27.65) \\
Unperturbed + Denoising & 0.76 (0.19) & 0.64 (0.22) & 0.76 (0.24) & 0.81 (0.14) & 0.95 (0.03) & 40.84 (28.00) \\
Unperturbed + Stochastic Ensemble & 0.76 (0.19) & 0.65 (0.22) & 0.78 (0.24) & 0.79 (0.15) & 0.95 (0.04) & 37.26 (28.25) \\
\midrule
SSAA (undefended) & 0.47 (0.27) & 0.35 (0.23) & 0.58 (0.33) & 0.52 (0.33) & 0.92 (0.04) & 47.64 (17.08) \\
SSAA + Randomized Preprocessing & 0.55 (0.23) & 0.42 (0.22) & 0.62 (0.29) & 0.62 (0.28) & 0.92 (0.04) & 48.31 (23.72) \\
SSAA + Denoising & 0.57 (0.22) & 0.43 (0.21) & 0.60 (0.28) & 0.66 (0.25) & 0.92 (0.04) & 49.55 (24.00) \\
SSAA + Stochastic Ensemble & 0.55 (0.24) & 0.41 (0.22) & 0.62 (0.30) & 0.61 (0.29) & 0.92 (0.04) & 49.10 (25.99) \\
\midrule
FDUA (undefended) & 0.65 (0.24) & 0.53 (0.24) & 0.71 (0.28) & 0.68 (0.25) & 0.94 (0.04) & 40.89 (25.54) \\
FDUA + Randomized Preprocessing & 0.67 (0.21) & 0.54 (0.23) & 0.68 (0.28) & 0.74 (0.19) & 0.94 (0.04) & 43.74 (27.24) \\
FDUA + Denoising & 0.67 (0.22) & 0.54 (0.24) & 0.67 (0.28) & 0.75 (0.20) & 0.94 (0.04) & 48.46 (27.78) \\
FDUA + Stochastic Ensemble & 0.66 (0.22) & 0.53 (0.22) & 0.69 (0.28) & 0.71 (0.21) & 0.94 (0.04) & 43.28 (25.96) \\
\bottomrule
\end{tabular}
\centering
\caption{Summary of segmentation metrics across all conditions. Values reported as mean (SD). DSC = Dice similarity coefficient; IoU = Intersection over Union; HD95 = 95th percentile Hausdorff distance; SSAA = Structured Speckle Amplification Attack; FDUA = Frequency-Domain Ultrasound Attack; SD = standard deviation.}
\label{tab:table1}
\end{table}

\subsection*{\textit{Attack Effectiveness}}

Figure 2 shows representative qualitative examples of segmentation outputs under unperturbed, adversarially attacked, and defended conditions for both SSAA and FDUA. SSAA reduced the mean DSC from 0.76 to 0.47 (SD 0.27), a mean DSC drop of 0.29 (95\% CI [0.22, 0.36], p $<$ 0.001, d = 1.46), while maintaining high visual similarity (SSIM = 0.94, L$_2$ = 5.9, L$_{\infty}$ = 0.27). The attack showed substantial variability, with individual clip DSC drops ranging from 0.04 to 0.78. Precision decreased from 0.81 to 0.58 (SD 0.33) and recall from 0.76 to 0.52 (SD 0.33), indicating that SSAA caused both missed nodule tissue (false negatives) and spurious predictions (false positives).

\begin{figure}[ht]
\centering
\includegraphics[width=0.75\linewidth]{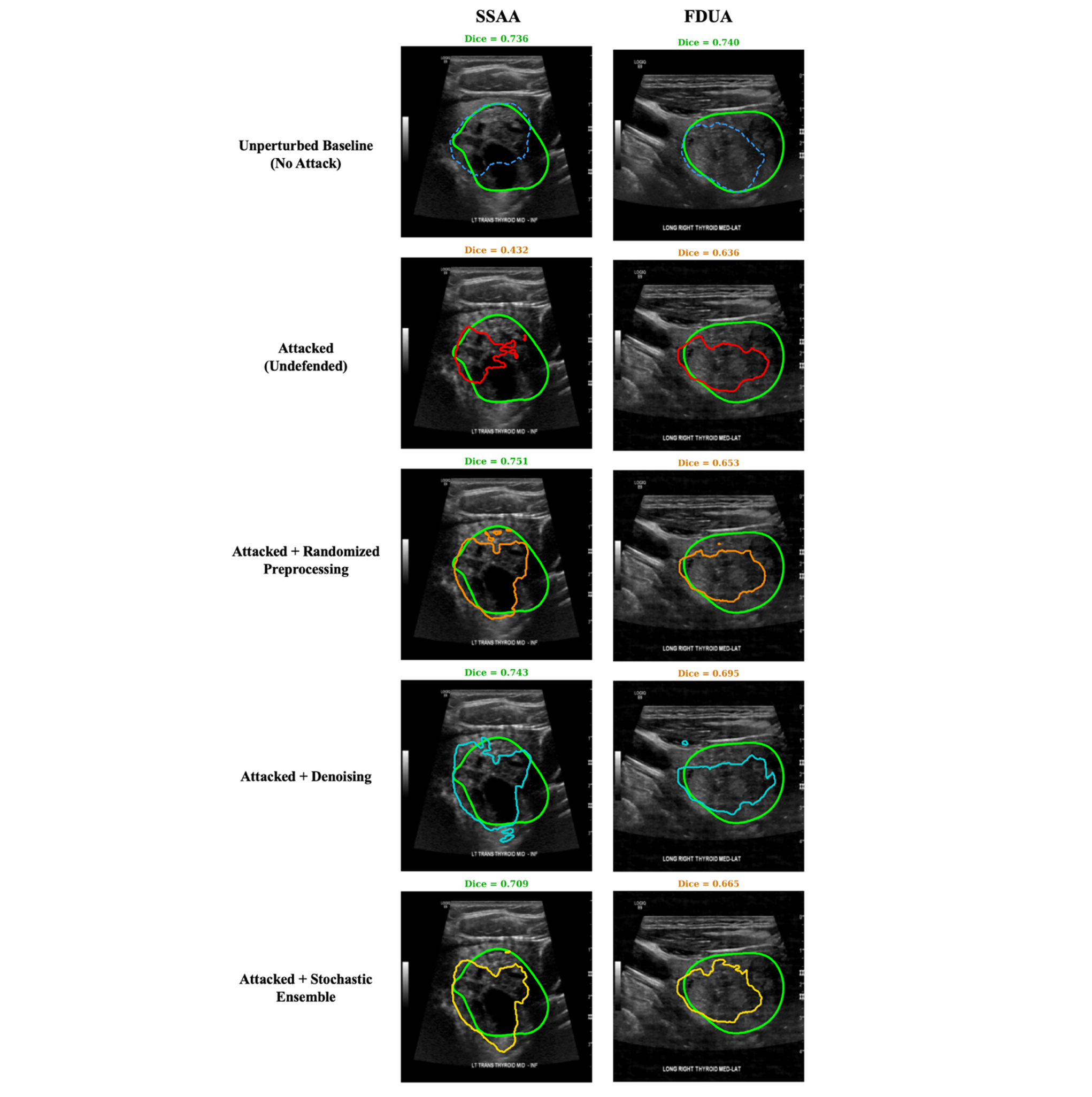}
\caption{Visual examples of segmentation from unperturbed to attacked to defended for representative test cases. Left image = Structured Speckle Amplification Attack (SSAA); Right image = Frequency-Domain Ultrasound Attack (FDUA); Green contour = ground truth; Blue dashed = unperturbed prediction; Red = undefended attacked prediction; Orange = randomized preprocessing; Cyan = deterministic denoising; Yellow = stochastic ensemble.}
\label{fig:fig2}
\end{figure}

FDUA produced a smaller mean DSC reduction from 0.76 to 0.65 (SD 0.24), a mean DSC drop of 0.11 (95\% CI [0.08, 0.14], p $<$ 0.001, d = 1.16), with lower visual fidelity (SSIM = 0.82, L$_2$ = 5.0, L$_{\infty}$ = 0.09). Precision was relatively preserved at 0.71 (SD 0.28) while recall dropped to 0.68 (SD 0.25), suggesting that FDUA primarily caused the model to under-segment the nodule rather than produce entirely erroneous predictions.

\subsection*{\textit{Unperturbed Performance Cost of Defenses}}

All three defenses imposed negligible cost on unperturbed image segmentation. Randomized preprocessing with test-time augmentation reduced unperturbed DSC by 0.0027, from 0.7601 to 0.7575. Deterministic input denoising reduced unperturbed DSC by 0.0031, from 0.7601 to 0.7570. The stochastic ensemble with consistency-aware aggregation produced a marginal improvement of 0.0005, from 0.7601 to 0.7606. Table 2 shows that all defenses imposed negligible performance cost on unperturbed images.

\begin{table}[ht]
\centering
\small
\begin{tabular}{lccc}
\toprule
\textbf{Condition} & \textbf{Unperturbed DSC} & \textbf{SD} & \textbf{DSC Cost} \\
\midrule
Undefended (baseline) & 0.7601 & 0.2036 & --- \\
Randomized Preprocessing & 0.7575 & 0.2027 & +0.0027 \\
Input Denoising & 0.7570 & 0.1850 & +0.0031 \\
Stochastic Ensemble & 0.7606 & 0.1863 & $-$0.0005 \\
\bottomrule
\end{tabular}
\caption{Unperturbed segmentation performance (no attack) with each defense, showing the cost of defense on unattacked images. DSC = Dice similarity coefficient; SD = standard deviation; DSC Cost = absolute change in Dice similarity coefficient relative to the undefended baseline.}
\label{tab:table2}
\end{table}

\subsection*{\textit{Randomized Preprocessing with Test-Time Augmentation}}

Against SSAA, randomized preprocessing significantly improved the mean DSC from 0.47 to 0.55, a gain of +0.09 (95\% CI [+0.04, +0.14], p = 0.0003, Cohen's d = 0.61). The DSC recovery rate was 29.29\%. Among secondary endpoints, IoU improved by +0.07 (p $<$ 0.001, d = 0.62), recall improved by +0.10 (p $<$ 0.001, d = 0.72, medium), precision improved by +0.04 (p = 0.14, d = 0.27), and pixel accuracy improved by +0.005 (p = 0.07, d = 0.27).

Against FDUA, randomized preprocessing did not significantly improve DSC (+0.016, 95\% CI [$-$0.03, +0.07], p = 0.36, d = 0.11). Despite the negligible DSC improvement, this defense significantly improved recall by +0.06 (p = 0.023, d = 0.36), but precision decreased by $-$0.03 (p = 0.89). This opposing recall-precision pattern explains why the net DSC effect was negligible. The DSC recovery rate was 16.92\%. Table 3 reports the statistical significance and effect sizes of DSC improvements for each defense--attack pairing.

\begin{table}[ht]
\centering
\small
\begin{tabular}{llcccc}
\toprule
\textbf{Attack} & \textbf{Defense} & \textbf{$\Delta$ DSC} & \textbf{95\% CI} & \textbf{p-value} & \textbf{Cohen's d} \\
\midrule
SSAA & Randomized Preprocessing & +0.09 & [+0.04, +0.14] & 0.0003 & +0.61 \\
SSAA & Denoising & +0.10 & [+0.06, +0.15] & 0.0002 & +0.73 \\
SSAA & Stochastic Ensemble & +0.08 & [+0.04, +0.12] & 0.0017 & +0.61 \\
FDUA & Randomized Preprocessing & +0.02 & [--0.03, +0.07] & 0.3603 & +0.11 \\
FDUA & Denoising & +0.01 & [--0.03, +0.05] & 0.1836 & +0.14 \\
FDUA & Stochastic Ensemble & +0.01 & [--0.02, +0.04] & 0.2647 & +0.10 \\
\bottomrule
\end{tabular}
\caption{Statistical significance of defense improvements on adversarial Dice similarity coefficient (DSC). Significance assessed by one-sided paired Wilcoxon signed-rank test with Bonferroni correction. SSAA = Structured Speckle Amplification Attack; FDUA = Frequency-Domain Ultrasound Attack (FDUA); CI = bootstrap 95\% confidence interval (1,000 resamples).}
\label{tab:table3}
\end{table}

\subsection*{\textit{Deterministic Input Denoising}}

Against SSAA, deterministic input denoising achieved the largest DSC improvement among all defenses, increasing the mean DSC from 0.47 to 0.57, a gain of +0.10 (95\% CI [+0.06, +0.15], p = 0.0002, Cohen's d = 0.73). The DSC recovery rate was 35.57\%. Among secondary endpoints, IoU improved by +0.09 (p $<$ 0.001, d = 0.76), and recall showed the largest improvement across all defense-attack combinations at +0.14 (p $<$ 0.0001, d = 0.92). Precision improved by +0.03 (p = 0.12, d = 0.22). Pixel accuracy improved by +0.006 (p = 0.046, d = 0.31).

Against FDUA, deterministic input denoising did not significantly improve DSC (+0.015, 95\% CI [$-$0.03, +0.05], p = 0.18, d = 0.14). However, this defense produced a significant recall improvement of +0.07 (p $<$ 0.001, d = 0.76), the strongest recall recovery among all defenses against FDUA, corresponding to a recall recovery rate of 87.65\%. Precision decreased by $-$0.04 (p = 0.99, d = $-$0.40), offsetting the recall gain in the DSC. The DSC recovery rate was 20.60\%.

\subsection*{\textit{Stochastic Ensemble with Consistency-Aware Aggregation}}

Against SSAA, the stochastic ensemble improved the mean DSC from 0.47 to 0.55, a gain of +0.08 (95\% CI [+0.04, +0.12], p = 0.0017, Cohen's d = 0.61). The DSC recovery rate was 28.24\%. Among secondary endpoints, IoU improved by +0.07 (p = 0.0012, d = 0.63), recall improved by +0.09 (p $<$ 0.001, d = 0.75), and precision improved by +0.04 (p = 0.13, d = 0.21). Pixel accuracy improved by +0.004 (p = 0.06, d = 0.19).

Against FDUA, the stochastic ensemble did not significantly improve DSC (+0.009, 95\% CI [$-$0.02, +0.04], p = 0.26, d = 0.10). Recall improved by +0.03 (p = 0.035, d = 0.42), while precision decreased by $-$0.02 (p = 0.82, d = 0.19). The DSC recovery rate was 8.06\%, the lowest among all defenses against FDUA. Figure 3 summarizes the effects of each defense across all attacks using multiple segmentation metrics, including DSC, IoU, precision, and recall.

\begin{figure}[ht]
\centering
\includegraphics[width=\textwidth]{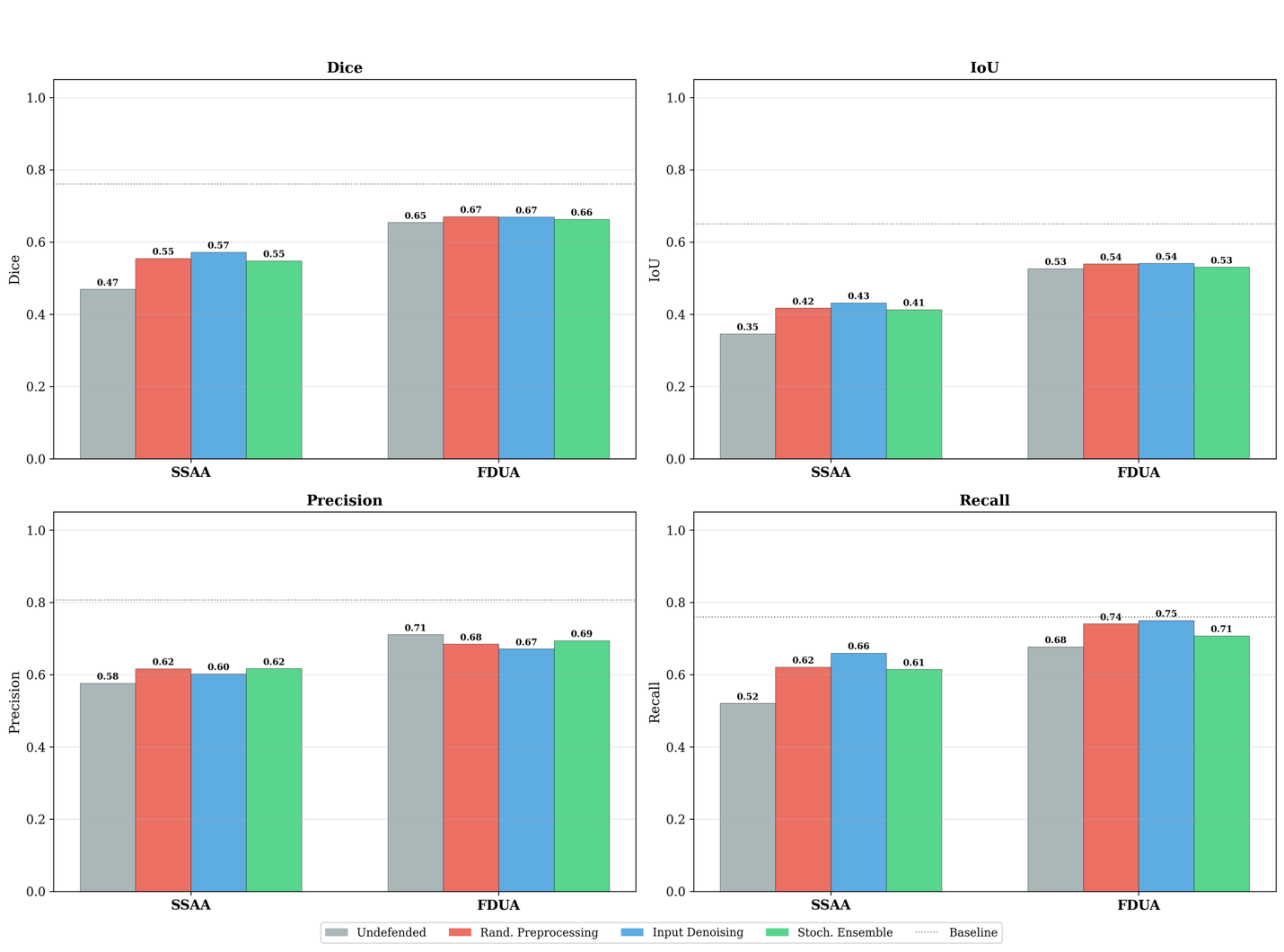}
\caption{Multi-metric comparison across all attack and defense conditions. Dice = Dice similarity coefficient; IoU = Intersection over Union; SSAA = Structured Speckle Amplification Attack; FDUA = Frequency-Domain Ultrasound Attack.}
\label{fig:fig3}
\end{figure}

\subsection*{\textit{Cross-Defense Comparison and Recovery Rates}}

Against SSAA, all three defenses achieved comparable DSC recovery (28.24\% to 35.57\%), with deterministic input denoising producing the highest recovery. IoU recovery closely paralleled DSC at 26.73\% to 33.48\%. The defenses primarily recovered missed nodule tissue rather than correcting spurious predictions, as recall recovery rates (26.52\% to 35.66\%) substantially exceeded precision recovery rates ($-$4.63\% to 8.45\%). Pixel accuracy recovery ranged from 15.29\% to 18.00\%.

Against FDUA, DSC recovery rates were lower and not statistically significant (8.06\% to 20.60\%). A pattern emerged in which recall recovery rates were disproportionately higher (23.72\% for the stochastic ensemble, 62.60\% for randomized preprocessing, and 87.65\% for deterministic denoising) while precision recovery was consistently negative ($-$19.97\% to $-$29.26\%).

\section{Discussion}

This study shows that deep learning models for thyroid ultrasound segmentation can be affected by black-box adversarial perturbations that preserve visual similarity, and that the performance of the evaluated inference-time defenses varies depending on the domain in which the attack operates. In particular, SSAA showed partial mitigation with standard preprocessing, whereas FDUA was not significantly mitigated by the tested defenses.

Even under a constrained query budget, the attacks were able to meaningfully degrade segmentation performance. For SSAA, the mean DSC decreased from 0.76 to 0.47, a reduction that could correspond to clinically relevant underestimation or displacement of a thyroid nodule. Importantly, these perturbations were visually subtle, making them possibly difficult to detect through routine image review. This highlights that black-box access alone may be sufficient to induce clinically meaningful model failure.

The observed difference in defensibility between SSAA and FDUA is consistent with prior work suggesting that defense effectiveness depends on the relationship between perturbation structure and image features used by the model \cite{ref20,ref21}. In ultrasound, where speckle and tissue texture define much of the signal content, spatially localized perturbations may be more amenable to smoothing or denoising than perturbations embedded within frequency components that overlap with normal texture patterns \cite{ref9,ref10,ref22}. While this study did not directly measure model feature sensitivity, the differential response to preprocessing supports this interpretation.

From a practical standpoint, a simple deterministic denoising pipeline recovered approximately 36\% of the lost DSC under SSAA with negligible cost to unperturbed performance. Randomized preprocessing and stochastic ensemble inference also produced meaningful but smaller recovery, restoring approximately 29\% and 28\% of lost DSC, respectively. These gains were driven primarily by improvements in recall, indicating partial recovery of missed nodule tissue rather than correction of spurious predictions.

In contrast, none of the tested defenses produced a statistically significant improvement in DSC under FDUA. One possible explanation is that FDUA introduces perturbations that are not easily separable from normal ultrasound appearance using simple input-level preprocessing. Qualitatively, defended images remained visually distinct from unperturbed images even after denoising, suggesting that the perturbations persist despite filtering. This interpretation is supported by the secondary metric patterns observed for FDUA, where recall consistently improved across defenses while precision decreased, reflecting a shift toward over-segmentation. These opposing effects largely cancelled in the DSC, resulting in minimal net improvement.

Together, these findings indicate that input-level preprocessing alone is unlikely to provide reliable protection against frequency-domain attacks in ultrasound segmentation. The resistance of FDUA to preprocessing, contrasted with the partial mitigation observed for spatial-domain attacks such as SSAA, highlights a clear asymmetry in defensibility. This reinforces the need to evaluate adversarial robustness in a modality- and attack-specific manner rather than assuming that a single class of defense will generalize across perturbation types \cite{ref5,ref8}.

Several directions for future work follow from this study. The limited effectiveness of preprocessing against FDUA motivates investigation of training-time strategies that expose models to frequency-domain perturbations, while the partial recovery achieved for SSAA suggests that incorporating speckle-like perturbations during training may further improve stability. Ensemble approaches using architecturally diverse models, rather than multiple augmented copies of a single network, may further reduce correlated failure modes and improve robustness under both spatial- and frequency-domain perturbations \cite{ref23,ref24}. Beyond mitigation, the observed consistency patterns across stochastic predictions point to a practical role for confidence-based triage mechanisms that flag unstable or low-consensus segmentations for human review \cite{ref5,ref7,ref25}. Such approaches may be particularly important in post-market deployment, where ongoing monitoring of model behavior and failure patterns is necessary to detect emerging vulnerabilities and support safe clinical use over time.

This study was not without its limitations. The evaluation was conducted on a single dataset with a single model architecture, and generalizability to other datasets, anatomical sites, or architectures cannot be assumed. In addition, the attacks used a fixed query budget of 500, and stronger attacks with larger budgets could potentially overcome the defenses that showed effectiveness. Lastly, the defenses were evaluated under a non-adaptive threat model, and adaptive attacks specifically designed to circumvent the tested defenses could be more effective.

\section{Conclusion}

This study demonstrates that deep learning models for thyroid nodule ultrasound segmentation are vulnerable to black-box adversarial perturbations. Spatial-domain attacks were partially mitigated by simple inference-time preprocessing, recovering up to 36\% of lost DSC without measurable cost to unperturbed performance, whereas frequency-domain attacks were not mitigated by any of the tested defenses. These results highlight a modality-specific asymmetry in defensibility. Future work should focus on training-level and architectural approaches to improve robustness, and on evaluating these strategies across additional ultrasound tasks to support safe and reliable clinical deployment.

\section*{Data Availability Statement}

The dataset used in this study is the publicly available Stanford AIMI Thyroid Ultrasound Cine-clip dataset \cite{ref18}, which contains biopsy-confirmed thyroid nodules with radiologist-annotated segmentation masks. The data is available at: \url{https://doi.org/10.71718/7m5n-rh16}.

\end{document}